# Fully Elman Neural Network: A Novel Deep Recurrent Neural Network Optimized by an Improved Harris Hawks Algorithm for Classification of Pulmonary Arterial Wedge Pressure

Masoud Fetanat*, Michael Stevens, Pankaj Jain, Christopher Hayward, Erik Meijering, *Fellow, IEEE,* Nigel H. Lovell, *Fellow, IEEE*

**Abstract**— Heart failure (HF) is one of the most prevalent life-threatening cardiovascular diseases in which 6.5 million people are suffering in the USA and more than 23 million worldwide. Mechanical circulatory support of HF patients can be achieved by implanting a left ventricular assist device (LVAD) into HF patients as a bridge to transplant, recovery or destination therapy and can be controlled by measurement of normal and abnormal pulmonary arterial wedge pressure (PAWP). While there are no commercial long-term implantable pressure sensors to measure PAWP, real-time non-invasive estimation of abnormal and normal PAWP becomes vital. In this work, first an improved Harris Hawks optimizer algorithm called HHO+ is presented and tested on 24 unimodal and multimodal benchmark functions. Second, a novel fully Elman neural network (FENN) is proposed to improve the classification performance. Finally, four novel 18-layer deep learning methods of convolutional neural networks (CNNs) with multi-layer perceptron (CNN-MLP), CNN with Elman neural networks (CNN-ENN), CNN with fully Elman neural networks (CNN-FENN), and CNN with fully Elman neural networks optimized by HHO+ algorithm (CNN-FENN-HHO+) for classification of abnormal and normal PAWP using estimated HVAD pump flow were developed and compared. The estimated pump flow was derived by a non-invasive method embedded into the commercial HVAD controller. The proposed methods are evaluated on an imbalanced clinical dataset using 5-fold cross-validation. The proposed CNN-FENN-HHO+ method outperforms the proposed CNN-MLP, CNN-ENN and CNN-FENN methods and improved the classification performance metrics across 5-fold cross-validation with an average sensitivity of 79%, accuracy of 78% and specificity of 76%. The proposed methods can reduce the likelihood of hazardous events like pulmonary congestion and ventricular suction for HF patients and notify identified abnormal cases to the hospital, clinician and cardiologist for emergency action, which can diminish the mortality rate of patients with HF.

*Masoud Fetanat (correspondence e-mail:m.fetanat@ieee.org), Michael Stevens and Nigel Lovell are with the Graduate School of Biomedical Engineering, UNSW, Sydney, Australia. Pankaj Jain and Christopher Hayward are with Cardiology Department, St Vincent's Hospital, Sydney, Australia; Victor Chang Cardiac Research Institute, Sydney, Australia; and School of Medicine, UNSW, Sydney, Australia. Erik Meijering is with the Graduate School of Biomedical Engineering and School of Computer Science and Engineering, UNSW, Sydney, Australia.

## I. INTRODUCTION

Cardiovascular disease (CVD) is a highly prevalent illness affecting 26.7 million people in the United States of America [1]. 6.5 million people are suffering from heart failure (HF) in the USA and more than 23 million worldwide [2]. The prevalence of HF will increase by 46% from 2012 to 2030, leading to more than 8 million adult with HF according to the American Heart Association [1]. Although heart transplantation is the gold standard treatment for end-stage HF patients, the number of donor hearts is significantly less than the demand for heart transplantation and therefore the mortality of people on the waitlist is high. Mechanical circulatory support is one of the most suitable treatments for HF patients, and is achieved by implanting a mechanical pump, called a left ventricular assist device (LVAD), into the body of the HF patients [1], [2].

Currently LVADs are operated at constant speed which makes them insensitive to changes in demand as physiological conditions change [3]. This can lead to hazardous events like ventricular suction (ventricular collapse because of low pressure in the ventricle) or pulmonary congestion (excess fluid in the lungs because of high pressure in the ventricle) [4]. Physiological control systems for LVADs can be used to automatically adjust pump speed to maintain a constant left ventricular end-diastolic pressure (LVEDP) [4], [5]. Maintaining this important clinical variable can supplement the Frank-Starling response of the native heart and therefore can prevent these hazardous events for HF patients [6]–[9].

This physiological control system approach requires an implantable pressure sensor to measure LVEDP in HF patients. However, there are currently no commercial long-term implantable pressure sensors available. An alternative is to use pulmonary arterial wedge pressure (PAWP) as it can be measured more easily as a surrogate index for LVEDP by catheterization [3], [4], [10], [11]. Nevertheless, measuring the PAWP by pulmonary artery catheterization can only work for a short period of time in hospital settings as most of these patients are discharged home, and this technique not being suitable of safe outside the hospital environment. Therefore, the non-invasive prediction of abnormal and normal PAWP becomes more critical. If these values can be predicted, a physiological control system is able to automatically adjust the pump speed to restore PAWP to a safe level.

Machine learning algorithms can be employed for identifying the abnormal and normal PAWP using a non-invasive signal from the implanted pump [11]. These algorithms mainly include feature extraction, feature selection and classification parts. However, manually extracting and selecting suitable features is time-consuming and can lead to overfitting [12]. Furthermore, extracting features manually can reduce the generalization ability of the machine learning methods [12].

To overcome this issue, deep learning methods such as convolutional neural networks (CNNs) can be used to automatically extract and select the features from the raw input signals or images [12]–[14]. Previously, a variety of deep learning methods have been used for the classification of cardiovascular disease using electrocardiogram (ECG) signals [15]–[27]. Furthermore, several previous studies have employed recurrent neural networks (RNN) for classification of CVDs [28], [29]. The results demonstrated that RNNs outperform static conventional neural networks like a multi-layer perceptron (MLP) as RNNs have a dynamic processing ability [12], [30]–[34]. However, the Elman neural network (ENN) is standing out among many classical types of RNNs and is widely applied in different research fields [31], [35]–[37]. An ENN, which has a dynamic memory ability, can use the previous and current features of the input signals for improving the classification performance metrics and is more suited for the small imbalanced dataset used in this work. Modification of ENN can further improve the classification performance.

CNNs suffer from the need for many hyper-parameters, which should be optimized during training of the networks. However, optimizing hyper-parameters manually is very challenging [38]–[40].

Moreover, the hyper-parameters are data dependent and therefore may not be appropriate in another dataset. Obtaining the appropriate values of the hyper-parameters of CNNs is not an easy task since there is no robust mathematical approach which can be employed. Hence, determining the values of the hyper-parameters of CNNs necessitates many iterations to achieve better performance [39]. Random and grid searches could be employed to determine the hyper-parameters automatically, however these are time consuming [39]. In addition, Bayesian optimization methods could be also used to achieve the hyper-parameters. However, it requires estimation of several statistics of the error function, which can lead to an inefficient results [41], [42]. Proposing an automatic method based on the metaheuristic algorithms for optimization of the values of these parameters can significantly result in less computational cost and higher performance [43]. These metaheuristic algorithms can also work on non-continuous, and non-differentiable problems [44], [45].

There are many metaheuristic algorithms such as Particle Swarm Optimization (PSO) [46], Differential Evolution (DE) [47], Gravitational Search Algorithm (GSA) [48], Harris Hawks Optimization (HHO) [44], Marine Predators Algorithm (MPA) [49], Slime Mould Algorithm (SMA) [50], Equilibrium Optimizer (EO) [51], Salp Swarm Algorithm (SSA) [52], Grey Wolf Optimizer (GWO) [53], Preaching Optimization Algorithm (POA) [54], Red fox optimization algorithm (RFO) [55], African Vultures Optimization Algorithm (AVOA) [56] and Artificial Gorilla Troops Optimizer (GTO) [57]. In metaheuristics, two major strategies of exploration and exploitation of random search are executed. While exploration refers to finding relevant areas in the solution space, exploitation concerns the extraction of useful information from these areas [58]. Based on the “no free lunch (NFL)” theorem [59], which states that there is no ideal metaheuristic to solve all types of optimization problems well [60], any given problem may benefit from an improved metaheuristic algorithm. The HHO algorithm outperformed many other metaheuristic algorithms for finding optimal solution in different benchmark functions [44], [60], [61], and several applications such as image thresholding problems [62], estimation of photovoltaic models [63] and drug design [64]. The HHO algorithm was inspired by chasing and escaping behavior between Harris Hawks and rabbits as was recently introduced by Heidari et al. [44]. Furthermore, Heidari et al. showed that HHO can fail in providing optimal solutions in some cases [44], [60]. These findings motivated us to try to improve the well-known HHO algorithm.

This is the first study on detection of abnormal and normal PAWP non-invasively from pump flow for HF patients with implanted LVADs using a deep learning method. This study aims to improve the performance of the abnormal and normal PAWP classifications which can be employed in real-time. Although there are many deep learning methods in the literature, none to date have considered the effect of a recurrent neural network in the last layers of a CNN for this purpose. The proposed network has connections between two consecutive time points for the output layer, two consecutive time points for the hidden layer, input layer to output layer, and output layer to hidden layer. The hypothesis of this study was that these connections may fully use the long-term, spatial and temporal information of the input, hidden and output variables. Furthermore, the conventional optimization methods in the literature might not be capable of escaping from the local optima. Therefore, a new optimization method was proposed not only to optimize the hyper-parameters of the proposed deep learning methods, but also to improve the performance over standard benchmark functions. The classification of PAWP based on the pump flow signal from the commercial HVAD controller can be used to adjust the pump speed in real-time and therefore reduce the likelihood of hazardous events like pulmonary congestion and ventricular suction. Furthermore, the proposed method can be implemented on a mobile software platform and notify identified abnormal cases to the hospital, clinician and cardiologist for emergency action, which can diminish the mortality rate patients with HF.

The novelty of this work is multifold, first a modified HHO algorithm called HHO+ is proposed. Second, a new fully Elman neural network called FENN is proposed, which employs long-term and temporal information of the input and output variables that can improve the classification performance. Third, the hyper-parameters of FENN were optimized by the proposed HHO+ algorithm to further improve the classification performance. Finally, four non-invasive approaches for the classification of abnormal and normal PAWP via combination of CNN, ENN, FENN and HHO+ using the pump flow signal are proposed and compared. Relevant features are automatically extracted from the input pump flow signal by a CNN.

In the following sections, first the clinical dataset and preprocessing are described. Next, the HHO, CNN and ENN algorithms are introduced. Then, an improved Harris Hawks algorithm (HHO+) and novel fully Elman neural network (FENN) are presented. The proposed HHO+ is compared with recent metaheuristics such as HHO [44], MPA [49], SMA [50], EO [51], SSA [52] and GWO [53]. Subsequently, four novel methods for classification of PAWP, namely CNN-MLP, CNN-ENN, CNN-FENN and CNN-FENN-HHO+, are fully proposed. Afterwards, the performance evaluation metrics, K-fold stratified cross validation procedure and weighted cross-entropy loss function are described. Then, the proposed HHO+ is validated on 24 unimodal and multimodal benchmark functions. Subsequently, the comparative classification results of CNN-MLP, CNN-ENN, CNN-FENN and CNN-FENN-HHO+ on 5-fold cross-validation using a clinical patient dataset are presented. Finally, a discussion of the results, study limitations and future work concludes the paper.

## II. METHODOLOGY

### A. Dataset

The clinical data set was achieved from 25 stable and ambulatory patients with implanted HVAD (HeartWare, Medtronic) pump. PAWP was measured using right heart catheterization (7.5-French double transducer Swan-Ganz CCO catheter; Edwards Lifesciences, Irvine, CA) into the pulmonary artery of the implanted patients. HVAD pump flow was estimated by a non-invasive method embedded into the commercial HVAD controller. PAWP and pump flow were recorded in different pump speeds and exercise workloads. The sampling rate of the pressure and flow measurements were 50 Hz. The experimental study was approved by St Vincent's hospital research ethics committee [11]. The data were used for a 5-fold cross-validation dataset for evaluation of the proposed methods. PAWP of less than 8 or greater than 16 mmHg was taken to be abnormal, and other values as a normal condition, based on the clinicians' suggestions from St Vincent hospital in Sydney, Australia. The recorded dataset is an imbalanced dataset which has 459 abnormal PAWP and 1338 normal PAWP samples.

### B. Data pre-processing

A Butterworth low-pass filter with stop frequency of 15 Hz was employed to remove the higher frequency noise and smooth the PAWP and pump flow signals. The pump flow signal was then segmented with a length of 6 s (300 samples). The mean PAWP was calculated for each segment from all the samples between two consecutive minima using the filtered PAWP. PAWP has a pulsatile waveform due to heart contractility, and inhalation and exhalation. This process is shown in Fig. 1.

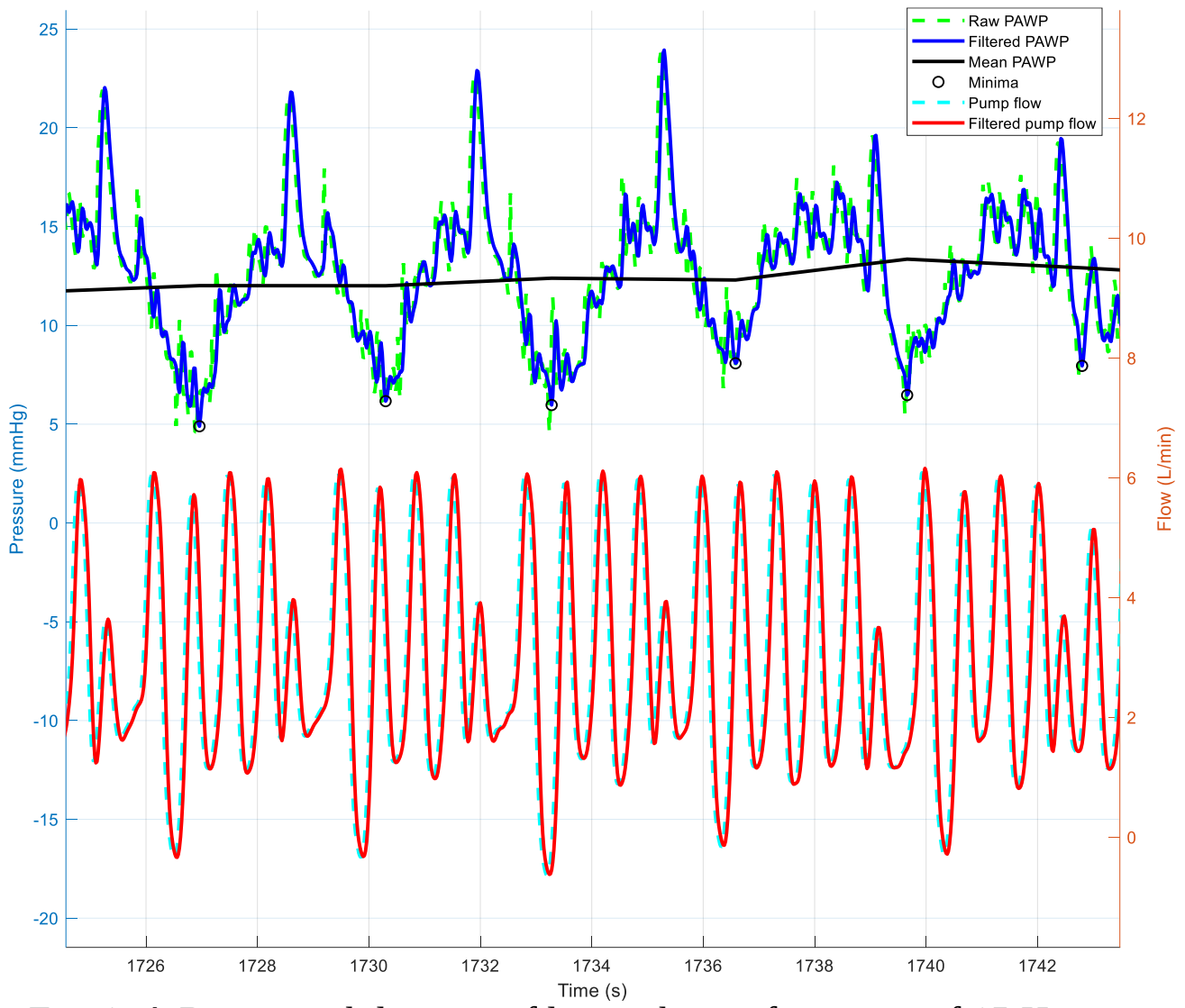

Fig. 1. A Butterworth low-pass filter with stop frequency of 15 Hz was employed to remove the higher frequency noise and smooth the PAWP and pump flow signals. The mean PAWP was calculated for each segment from all the samples between two consecutive minima using the filtered PAWP. PAWP: pulmonary arterial wedge pressure.

### C. Background materials

The HHO algorithm is a population-based and gradient-free optimization algorithm, which was inspired by exploring a prey (rabbit), surprising pounce, collaborative behavior of hawks, and different attacking strategies for catching a prey. Although the hawks first look for different potential locations to hunt, the rabbit exhibits escaping behavior to improve its survival chance. In this analogy, the rabbit represents the best solution, and the hawks represent candidate solutions in each iteration [44].

The CNN is one of the most popular deep learning methods and was proposed by Fukushima in 1980 and improved by Lecun in 1998 [65], [66]. CNNs can automatically extract features from the raw input signals using some filters and then map the features to the labeled classes [31].

ENNs are similar to feed-forward neural networks, except that the hidden layer has a recurrent connection with a certain delay to each neuron, allowing the network a dynamic response to time dependent input data, which can increase the performance of the classification [12].

Due to space limitations on the manuscript, the HHO, CNN and ENN algorithms are described in detail in supplementary materials in sections 1, 2 and 3, respectively.

### D. Improved Harris Hawks algorithm (HHO+)

#### 1) Improved exploration and exploitation algorithm (IEEA)

In the conventional HHO algorithm, the exploration and exploitation phases were only defined based on the escaping energy of the rabbit. However, this cannot give the whole required information for different attacking strategies on catching a prey as other factors such as velocity of the hawks and rabbit are also very important in each step of the HHO algorithm.

The marine predator algorithm (MPA) [49] is a new metaheuristic algorithm mimicking the whole life of a predator and prey in terms of their velocities. The results showed that MPA is extensively able to explore and exploit the solution space [49]. However, the solutions derived from MPA can be trapped in local optima. Therefore, a new algorithm to improve the global solution and avoid trapping in local optima is presented here, which was inspired from MPA. In the improved exploration and exploitation algorithm (IEEA), some phases of MPA were modified and then combined, and an adaptive decreasing threshold was added to the algorithm, in which all have been determined experimentally. This algorithm mimics the process of hawks hunting a rabbit and the rabbit escaping in terms of their velocities and rapid movements. Accordingly, employing the IEEA can prevent the solutions being trapped into local optima and improve the optimal solution.

The improved exploration and exploitation algorithm includes two sections: (1) when the rabbit is escaping equal to or faster than the hawks are moving and (2) when the hawks are moving faster than the escaping rabbit. The first phase occurs in the initial iteration of optimization, where exploration is more important than exploitation. The second phase happens in the intermediate and last phases of optimization, where the exploration attempts are converted to exploitation gradually or exploitation matters more than exploration.

These two phases are mathematically described as follows:

$$X_{t+1} = \begin{cases} X_t + rand \times R_G \times (X_{rabbit,t} - R_G \times X_t) & if\ rand < a \\ (X_{rabbit,t} + rand \times R_G \times (R_G \times X_{rabbit,t} - X_t) & otherwise \end{cases} \quad (1)$$

where $\mathrm{X}_{t+1}$ and $X_t$ are the position vectors of the hawks in iteration $t+1$ and $t$, respectively. $X_{rabbit,t}$ denotes the position of the rabbit (best solution) in iteration $t$. $a$ is an adaptive decreasing threshold for switching between exploration and exploitation algorithms. $rand$ denotes a uniformly distributed random number in the interval (0,1) and $R_G$ is a vector including random numbers based on the zero-mean and unit-variance Gaussian distribution:

$$R_G(x) = \frac{1}{\sqrt{2\pi\sigma^2}} exp\left(-\frac{(x-\mu)^2}{2\sigma^2}\right) = \frac{1}{\sqrt{2\pi}} exp\left(-\frac{x^2}{2}\right) \quad (2)$$

$\mu$ and $\sigma$ denote the mean and variance of the Gaussian distribution, respectively.

The adaptive decreasing threshold $a$ is defined as follows:

$$a = tanh(-\left(\frac{t}{T}\right) + 1) \quad (3)$$

where $t$ and $T$ are the current and maximum number of iterations, respectively. A maximum number of iterations is employed as the stopping criterion for the HHO+ algorithm. To compare Levy-flight and Gaussian distributions, 2D trajectory of Levy-flight and Gaussian distributions are shown in Fig. S4a and 4b, respectively.

*2) Quasi-oppositional based learning (QOBL)*

Oppositional based learning (OBL) was originally proposed by Tizhoosh [67], which put forward the concept of an opposite point. Tizhoosh showed that opposite numbers can improve searching ability, solution accuracy and convergence speed of population-based optimization algorithms compared with random numbers [67]. Rahnamayan et al. [68] proposed a new OBL method called quasi-oppositional based learning, which can further improve the solution accuracy for finding the global optimal points, as follows.

Let $X(x_1, x_2, \dots, x_n)$ be a point, $X^0(x_1^0, x_2^0, \dots, x_n^0)$ an opposite point, and $X^{q0}(x_1^{q0}, x_2^{q0}, \dots, x_n^{q0})$ the quasi-opposite point, all in n-dimensional space. $x_i^0$ and $x_i^{q0}$ can be defined by as:

$$x_i^0 = a_i + b_i - x_i \quad (4)$$

where $x_i \in R$ and $x_i \in [a_i, b_i], \forall i \in 1, 2, \dots, n$.

$$x_i^{q0} = rand(\frac{a_i + b_i}{2}, x_i^0) \quad (5)$$

If the current solution is too far from the optimal solution, then computing the quasi-oppositional location can result in the opposite direction, which can get closer to the optimal solution. Therefore, employing QOBL algorithm can improve the exploration ability of population-based optimization algorithms and prevent the solutions falling into local optima [69]. The pseudo code of (QOBL) is shown in Algorithm S2.

*3) Proposed HHO+*

The proposed HHO+ is made by combining HHO, IEEA, and QOBL. The pseudo code of HHO+ is given in Algorithm 1. As shown in the pseudo code, first hawk populations are randomly generated. Afterward, each hawk's fitness is calculated, and the best location of hawks is stored in $X_{rabbit}$ same as the classical HHO. Then, the locations of hawks are updated based on the values of $E$ and $r$ using Eqs. (S4), (S6), (S10) and (S11). Afterward, IEEA and QOBL are executed and $X_{rabbit}$ is updated. Finally, the algorithm returns $X_{rabbit}$ if the current iteration exceeds the maximum number of iterations, otherwise the process is repeated.

In the experiments, the proposed HHO+ is compared with some well-known and recent metaheuristic algorithms such as HHO [44], MPA [49], SMA [50], EO [51], SSA [52] and GWO [53]. The average results of the proposed HHO+ and other algorithms are ranked by the non-parametric statistical Friedman and Wilcoxon rank sum test.

## *E. Proposed fully Elman neural network (FENN)*

In the standard ENN, there is only one feedback between the context layer and hidden layer at time $t$. This feedback is not able to fully use the long-term, spatial and temporal information of the input and output variables [31], [37]. Accordingly, to improve the performance of the ENN, a fully Elman neural network structure is created in this work, which has connections between two consecutive time points for the output layer, two consecutive time points for the hidden layer, input layer to output layer, and output layer to hidden layer.

Fig. 2 depicts the architecture of the proposed FENN model. The mathematical model of the FENN is presented as follows:

$$y(t) = f(w_2\, x(t) + w_3\, u(t-1) + y_{c2}(t) + b_2) \quad (6)$$

$$x(t) = g(x_c(t) + w_1\, u(t-1) + y_{c1}(t) + b_1) \quad (7)$$

$$x_c(t) = w_4\, x_c(t-1) + w_5\, x(t-1) \quad (8)$$

$$y_{c1}(t) = w_6\, y_{c1}(t-1) + w_7\, y(t-1) \quad (9)$$

$$y_{c2}(t) = w_8\, y_{c2}(t-1) + w_9\, y(t-1) \quad (10)$$

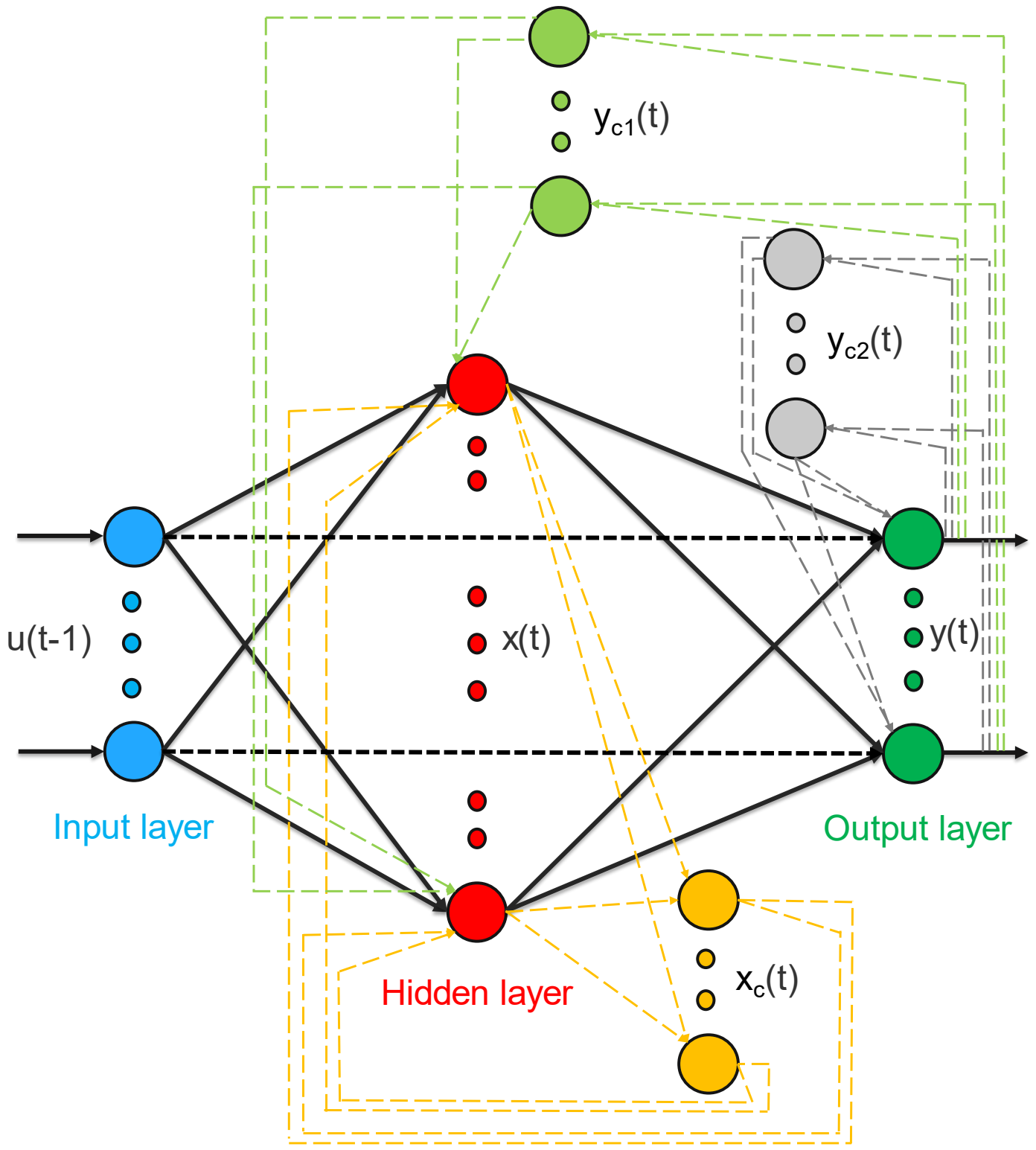

Fig. 2. Architecture of the proposed FENN model. u(t-1) is the vector of the input layer at the moment t-1, $\boldsymbol{x(t)}$ represents the vector of hidden layer, and $\boldsymbol{x_c(t)}, \boldsymbol{y_{c1}(t)}$ and $\boldsymbol{y_{c2}(t)}$ represent context layer for hidden layer, output context layer 1 and output context layer 2 at the moment $\boldsymbol{t}$, and $\boldsymbol{y(t)}$ is the vector of output layer at the time t.

**Algorithm 1** Pseudo code of the proposed HHO+ algorithm

**Inputs:** The population size N and maximum number of iterations T
Initialize the random hawk populations $X_i(i = 1, 2, \ldots, N)$ in the specific search space
**while** (maximum iteration is not met) **do**
  Calculate the fitness values of hawks
  Set $X_{rabbit}$ as best solution of hawks
  **for** (each hawk ($X_i$)) **do**
    Update the initial energy E0 and jump strength J
    Update the E using Eq. (S3)
    **If** ($|E| \geq 1$) **then** Exploration phase
      Update the hawk vectors using Eq. (S1)
    **else if** ($|E| < 1$) **then** Exploitation phase
      **if** ($r \geq 0.5$ and $|E| \geq 0.5$) **then** Soft besiege
        Update the hawk vectors using Eq. (S4)
      **else if** ($r \geq 0.5$ and $|E| < 0.5$) **then** Hard besiege
        Update the hawk vectors using Eq. (S6)
      **else if** ($r < 0.5$ and $|E| \geq 0.5$) **then** SBWPRD
        Update the hawk vectors using Eq. (S10)
      **else if** ($r < 0.5$ and $|E| < 0.5$) **then** HBWPRD
        Update the hawk vectors using Eq. (S11)
      **end**
    **end**
    Update $X_{rabbit}$ by the proposed IEEA using Eqs. (1), (2) and (3)
    Apply QOBL in Algorithm S2 and update $X_{rabbit}$
  **end**
**end**
**Return** The location of rabbit ($X_{rabbit}$) and its fitness value where $w_2$, $w_3$ and $w_1$ are the connection weights between hidden layer to output layer, input layer to hidden layer, and input layer to hidden layer, respectively. $y_{c1}(t)$, $y_{c2}(t)$ and $x_c(t)$ denote the output context layer 1, output context layer 2 and context layer for hidden layer at $t$, respectively. $b_2$ and $b_1$ are the bias for output layer and hidden layer. $w_4$, $w_6$ and $w_8$ represent the recurrent connection weights between the moment $t$ and $t-1$ of $x_c$, $y_{c1}$ and $y_{c2}$, respectively.

$w_5$, $w_7$ and $w_9$ indicate the connection weights of $x(t-1)$ to $x_c(t)$, $y(t-1)$ to $y_{c1}(t)$, and $y(t-1)$ to $y_{c2}(t)$. $y(t)$ and $x(t)$ represent the vector of the output layer and hidden layer at $t$, respectively, and $u(t-1)$ is the vector of the input layer at $t-1$. $f$ and $g$ represent the activation function of the output layer and hidden layer, which were chosen as softmax and hyperbolic tangent sigmoid, respectively.

### *F. Proposed combined CNN with FENN and HHO+ (CNN-FENN and CNN-FENN-HHO+)*

In this study, a CNN with FENN (CNN-FENN) and CNN with FENN optimized by HHO+ (CNN-FENN-HHO+) are proposed and compared to CNN with MLP (CNN-MLP) for classification of abnormal and normal PAWP using the features from the pump flow signal. All three CNNs are trained and validated with the same patient dataset.

In standard CNNs, a MLP is employed in the final layers to map the features to the labeled classes. However, the combination of CNN with MLP may result in poor classification performance on time-related data [12]. Therefore, in this study, the CNN-MLP in which the last two layers (fully-connected layers) are substituted with a FENN (CNN-FENN), is trained by the Adam algorithm [70] to classify the abnormal and normal PAWP. Fig. 3 depicts the architecture of the proposed CNN-FENN model.

The proposed deep learning methods were optimized by testing different filter sizes, number of layers and the types of each layer to minimize the loss function using grid search. To further improve the classification performance of CNN-FENN, three hyper-parameters related to training of CNNs, namely initial learning rate (ILR), learning rate drop factor (LRDF), and dropout probability (DP), were optimized by the proposed HHO+. Fig. 4 shows the flowchart of the proposed CNN-FENN-HHO+ system for classification of the abnormal and normal PAWP. In this hybrid algorithm, first the hawk populations are randomly generated, as in the classical HHO algorithm. Then, the fitness function in Eq. (14) is calculated by training the CNN-FENN and taking the average of the 10 results of training datasets. Afterward, the positions of hawks are updated using the proposed Algorithm 1. Once the current iteration reaches the maximum number of iterations, the found optimal values of ILR, LRDF and DP are fed to the CNN-FENN for validation on the test dataset. The final result is obtained by taking the average of the 5 results of the test dataset through 5-fold cross validation using the optimal parameters found by the proposed HHO+ algorithm.

### *G. Experimental setup, structure and parameters of HHO+, CNN-FENN and CNN-FENN-HHO+*

Table S2 shows the parameter settings for the optimization algorithms: HHO [44], MPA [49], SMA [50], EO [51], SSA [52] and GWO [53]. The proposed HHO+ is compared with these algorithms on 24 unimodal and multimodal benchmark optimization functions [53], [71] with the dimension of 1000. Table S3 shows the mathematical equations of these benchmark functions, depicted in Fig. S5. The population size for all the optimization algorithms and maximum iterations were set to 30 and 500, respectively [44]. All the optimization algorithms were randomly initialized. To increase the robustness of the result due to used random values in the optimization algorithms, each algorithm is initialized and run 30 times.

Table S4 represents the type of each layer, kernel sizes, activation functions and other parameters of the proposed CNN-FENN and CNN-FENN-HHO+, respectively. Layers 1 to 8 were created by convolution and batch normalization layers with LeakyReLU activation. Afterwards, layers 9 to 12 were constructed by convolution layers with LeakyReLU activation, max-pooling layers with stride 2 and batch normalization layers. Then, layers 13 to 16 were made by convolution and batch normalization layers with LeakyReLU activation. Finally, a ENN and FENN with 20 neurons in the hidden layer and activation function of hyperbolic tangent sigmoid (Tansig) and two neurons in the output layer with softmax activation were employed in layer 17 and 18 of CNN-FENN and CNN-FENN-HHO+, respectively.

As can be seen from Table S1 and Table S4 the same convolutional layers, batch normalization layers, size of kernels and activation functions were used for CNN-MLP, CNN-FENN and CNN-FENN-HHO+. The main difference among these methods is the last two layers which are used for classification

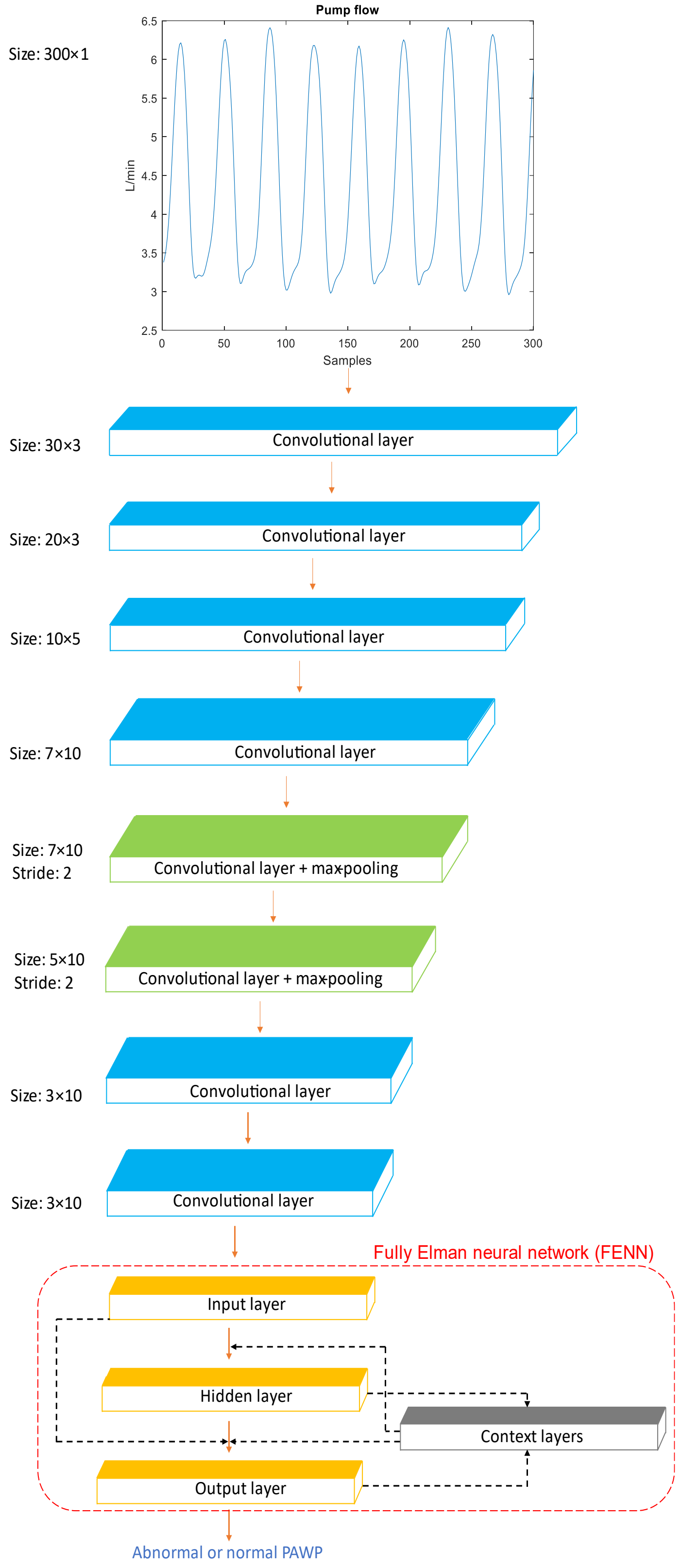

Fig. 3. Architecture of the proposed CNN-FENN model for classification of the abnormal and normal PAWP. 300 samples of pump flow signal are fed to the CNN-FENN as input. This is followed by three convolutional layers as well as batch normalization layers. Five convolutional, max-pooling layers with stride 2 and batch normalization layers are placed in the subsequent layers. The proposed FENN layer is located in the last layer of the CNN-FENN model to classify the abnormal and normal PAWP.

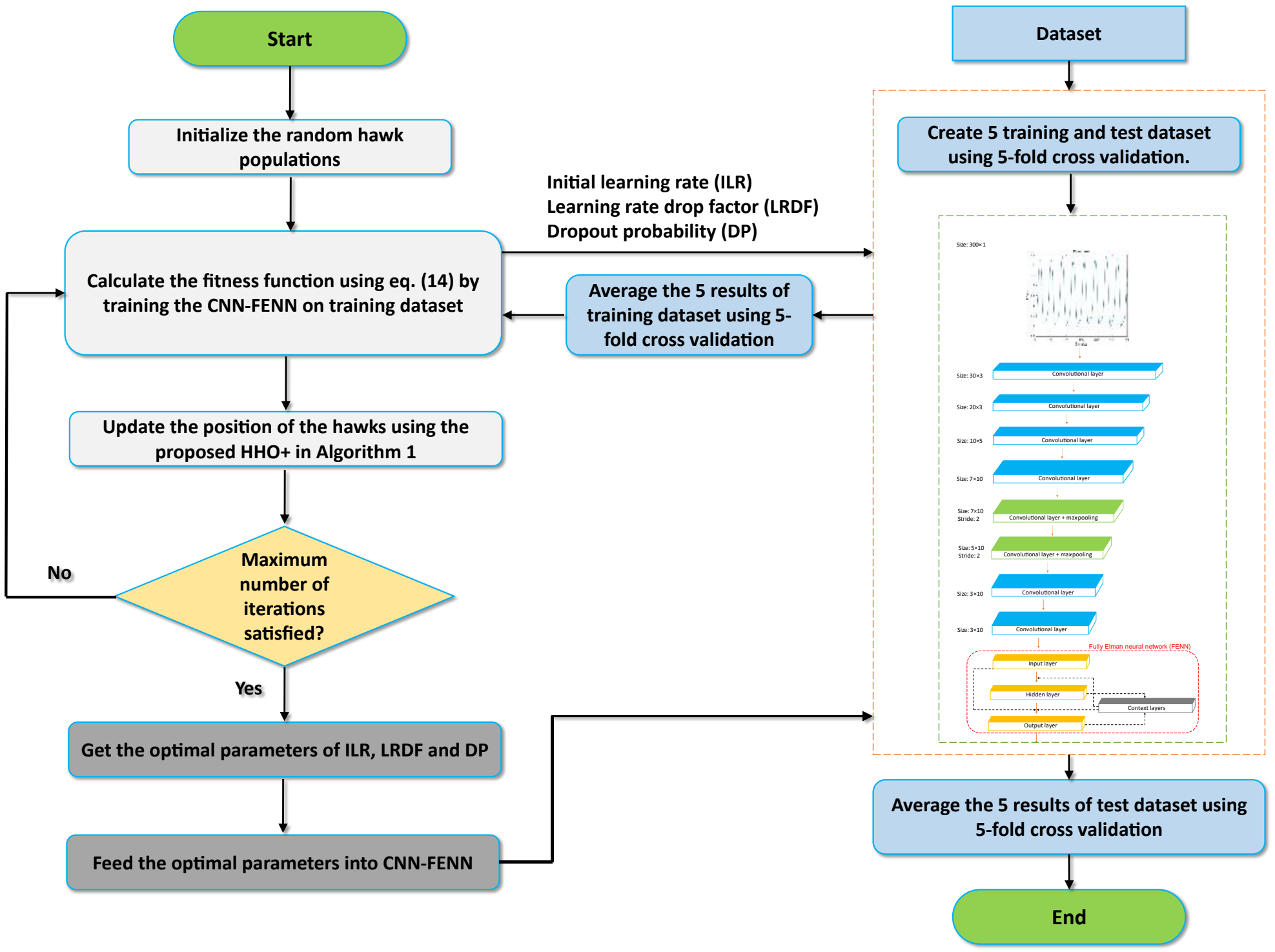

Fig. 4. Flowchart of the proposed CNN-FENN-HHO+ system.

of the features automatically extracted by the convolutional kernels.

The Adam algorithm was employed to optimize the weights and biases of the proposed CNN-MLP, CNN-FENN and CNN-FENN-HHO+ methods [70]. The algorithm by He was used to initialize the weights for CNN-MLP, CNN-FENN and CNN-FENN-HHO+ as it outperforms the well-known Xavier algorithm [72]. Furthermore, the proposed CNN-MLP, CNN-FENN and CNN-FENN-HHO+ were trained with 1000 iterations and a batch size of 179. This batch size value was set to one tenth of the total number of samples used in this study. All methods were implemented and run in MATLAB version 2020b (MathWorks, USA) and the experiments were performed on a computer with 16 GB of RAM, Core i7-8700K CPU@3.70 GHz.

### H. Performance evaluation metrics

Three standard performance metrics including accuracy, sensitivity and specificity were used to assess and evaluate the proposed methods:

$$Accuracy = \frac{TP + TN}{TP + FN + TN + FP} \quad (11)$$

$$Sensitivity = \frac{TP}{TP + FN} \quad (12)$$

$$Specificity = \frac{TN}{TN + FP} \quad (13)$$

where true positive (TP) denotes the number of abnormal PAWP classified correctly, true negative (TN) the number of the normal PAWP classified correctly, false positive (FP) indicates the number of normal PAWP classified incorrectly as abnormal PAWP, and false negative (FN) the number of the abnormal PAWP classified incorrectly as normal PAWP.

Accuracy refers to the proposed methods' ability to identify both abnormal and normal PAWP correctly, sensitivity is the proposed methods' ability to correctly detect abnormal PAWP, and specificity refers to the proposed methods' ability to identify the normal PAWP correctly.

### I. K-fold stratified cross validation

A 5-fold stratified cross validation [23] was employed to evaluate the ability of the proposed methods to classify abnormal and normal PAWP. Stratified cross validation was chosen due to the imbalanced dataset. It selects an equal portion of abnormal and normal PAWP for each fold of the dataset. First, the abnormal and normal PAWP samples were partitioned into ten parts, and nine parts of abnormal and normal PAWP samples were used to train the proposed methods and the rest were used for testing. Then, this process was repeated nine times to consider each of the ten folds for testing and the remaining folds for training. Moreover, 10% of the training dataset was assigned into a validation subset through cross-validation. In each fold, five patients which were not seen in the training dataset, were used in the test dataset for each iteration.

### J. Loss function

The classical cross-entropy function is not a suitable loss function due to the imbalanced clinical dataset used in this work. This function may result in high accuracy but very low sensitivity or specificity or will bias toward the majority class. Although it is important that the proposed methods correctly detect abnormal PAWP (high sensitivity), they should also be able to correctly identify normal PAWP (high specificity). The following weighted cross-entropy loss function [73] was employed to train the proposed deep neural networks for classifying abnormal and normal PAWP:

$$Loss\ Function = -\sum_{j=1}^{K} w_j \times T_j \times log(p_j) \qquad (14)$$

$$w_j = 1 - \frac{number\ of\ samples\ in\ the\ j^{th} class}{number\ of\ total\ samples}, \sum_{j=1}^{K} w_j = 1 \qquad (15)$$

where $w_j$ is the weight assigned to the class $j$, $K$ is the number of classes, $T_j$ is the $j^{th}$ element of the ground truth (target) vector and $p_j$ is the $j^{th}$ element of the estimated vector.

## III. Results

### A. Benchmark function validation

The comparative results of the proposed HHO+ on 24 unimodal and multimodal benchmark optimization functions for dimension 1000 are presented in Table S5.

The mean and standard deviation of 30 runs of each HHO+, HHO, MPA, SMA, EO, SSA and GWO algorithm are shown. According to this table, HHO+ achieved the best solutions on all the unimodal and multimodal benchmark optimization functions. As can be observed, the proposed HHO+ achieved an optimal fitness of 0 within 500 iterations with average and standard deviation of 0 in the benchmark functions of F1, F2, F7, F8, F11, F13, F21, F22, F23, and F24.

TABLE I shows the comparative results of two non-parametric statistical tests; Wilcoxon rank sum test [74] and Friedman's test [75] for dimension 1000. Wilcoxon rank sum test is a non-parametric statistical test that can determine the significant differences between the results of two algorithms. Friedman's test is able to rank the optimization methods based on their overall performance across different benchmark functions. According to this table, HHO+ obtained the best average ranking followed by HHO, SMA, MPA, EO, GWO and SSA, respectively. In the table, the signs '+', '=', and '-' indicate that the proposed HHO+ is significantly better than, equal or worse than the other optimization algorithms for significance level of 5%, respectively using Wilcoxon rank sum test. Friedman's test also showed the overall rank of the optimization methods on all the benchmark functions. The best results are marked in bold in Table S5 and TABLE I. It can be seen from TABLE I that HHO+ outperformed HHO, MPA, SMA, EO, SSA and GWO in 12, 22, 18, 24, 24, 24 benchmark functions, respectively, for a p-value of less than 0.05. Although the Wilcoxon rank sum test showed that HHO+ achieved better results than the other optimization methods for the majority of the benchmark functions, this merit was not statistically significant for the rest of the functions. Friedman's test results also showed that HHO+ was placed in the first rank compared to other competitors.

Fig. S6 shows convergence curves for F1 to F24, respectively. As can be seen from these figures, the HHO+ convergence curve is better (converges faster) than other optimization algorithms for both unimodal and multimodal benchmark optimization functions, except F11, F12 and F13, in which the HHO+ is ranked in second place.

Fig. S7 depicts the scalability results of the proposed HHO+ with different dimensions from 100 to 5000 in logarithm scale for the selected benchmark functions of F5, F6, F9, F15, F16 and F19 as the most challenging (not very close to the global minimum) functions shown in Table S3. Fig. S8 also shows the scalability results of the proposed HHO+ with different dimensions for F16 and F19 in normal scale to show the superiority of HHO+ more clearly. It can be observed from these figures that HHO+ outperforms the other optimization algorithms in both lower and higher dimensional problems.

Table S6 and S7 show the computational time of the optimization methods (HHO+, HHO, MPA, SMA, EO, SSA and GWO) on each of the 24 benchmark functions for dimension 1000 across 30 runs and the proposed deep learning methods (CNN-MLP, CNN-ENN, CNN-FENN and CNN-FENN-HHO+) across 30 training runs (10 epochs), respectively. It can be observed from these tables that the HHO optimization algorithm and CNN+MLP network structure have the minimum computational complexity.

TABLE I
STATISTICAL RESULTS OF WILCOXON RANK SUM TEST AND FRIEDMAN'S TEST FOR DIMENSION 1000. NAN REFERS TO NOT A NUMBER SHOWING ALL THE ELEMENTS OF TWO COMPARED VECTORS USING WILCOXON RANK SUM TEST ARE EQUAL TO ZERO.

| Fun. | HHO+ | HHO | MPA | SMA | EO | SSA | GWO |
|---|---|---|---|---|---|---|---|
| F1 | - | 1.21e-12 | 1.21e-12 | 1.27e-05 | 1.21e-12 | 1.21e-12 | 1.21e-12 |
| F2 | - | 3.02e-11 | 3.02e-11 | 3.02e-11 | 3.02e-11 | 3.02e-11 | 3.02e-11 |
| F3 | - | 1.21e-12 | 1.21e-12 | 6.25e-10 | 1.21e-12 | 1.21e-12 | 1.21e-12 |
| F4 | - | 3.02e-11 | 3.02e-11 | 5.57e-10 | 3.02e-11 | 3.02e-11 | 3.02e-11 |
| F5 | - | 7.96e-01 | 3.02e-11 | 9.92e-11 | 3.02e-11 | 3.02e-11 | 3.02e-11 |
| F6 | - | 7.28e-01 | 3.02e-11 | 8.10e-10 | 3.02e-11 | 3.02e-11 | 3.02e-11 |
| F7 | - | 1.21e-12 | 1.21e-12 | 2.93e-05 | 1.21e-12 | 1.21e-12 | 1.21e-12 |
| F8 | - | 1.21e-12 | 1.21e-12 | 3.34e-01 | 1.21e-12 | 1.21e-12 | 1.21e-12 |
| F9 | - | 9.03e-04 | 3.02e-11 | 6.70e-11 | 3.02e-11 | 3.02e-11 | 3.02e-11 |
| F10 | - | 1.26e-01 | 3.02e-11 | 1.44e-03 | 3.02e-11 | 3.02e-11 | 3.02e-11 |
| F11 | - | NaN | NaN | NaN | 5.51e-03 | 1.21e-12 | 1.21e-12 |
| F12 | - | NaN | 1.21e-12 | NaN | 1.21e-12 | 1.21e-12 | 1.21e-12 |
| F13 | - | NaN | 2.85e-04 | NaN | 4.17e-13 | 1.21e-12 | 1.21e-12 |
| F14 | - | 5.20e-01 | 3.02e-11 | 3.82e-10 | 3.02e-11 | 3.02e-11 | 3.02e-11 |
| F15 | - | 7.28e-01 | 3.02e-11 | 3.02e-11 | 3.02e-11 | 3.02e-11 | 3.02e-11 |
| F16 | - | 1.05e-01 | 3.02e-11 | 1.86e-09 | 3.02e-11 | 3.02e-11 | 3.02e-11 |
| F17 | - | 3.02e-11 | 3.02e-11 | 3.02e-11 | 3.02e-11 | 3.02e-11 | 3.02e-11 |
| F18 | - | NaN | NaN | 1.21e-12 | 1.21e-12 | 1.21e-12 | 1.21e-12 |
| F19 | - | 3.50e-09 | 1.86e-06 | 2.61e-10 | 4.08e-11 | 4.08e-11 | 3.02e-11 |
| F20 | - | 4.04e-01 | 3.02e-11 | 5.49e-01 | 3.02e-11 | 3.02e-11 | 3.02e-11 |
| F21 | - | 1.21e-12 | 1.21e-12 | 1.37e-03 | 1.21e-12 | 1.21e-12 | 1.21e-12 |
| F22 | - | 1.21e-12 | 1.21e-12 | 4.57e-12 | 1.21e-12 | 1.21e-12 | 1.21e-12 |
| F23 | - | 1.21e-12 | 1.06e-12 | 1.21e-12 | 9.56e-13 | 1.21e-12 | 1.21e-12 |
| F24 | - | NaN | 8.93e-13 | NaN | 1.14e-12 | 1.19e-12 | 1.21e-12 |
| Winner (+) / equal (=) / loser (-) | - | 12/12/0 | 22/2/0 | 18/6/0 | 24/0/0 | 24/0/0 | 24/0/0 |
| Friedman mean rank | **1.52** | 2.41 | 4.11 | 2.64 | 4.70 | 6.62 | 5.98 |
| Rank | **1** | 2 | 4 | 3 | 5 | 7 | 6 |

TABLE II
PERFORMANCE COMPARISON OF THE PROPOSED CLASSIFICATION METHODS ACROSS 5-FOLD CROSS-VALIDATION. THE NUMBERS ARE AVERAGES (STANDARD DEVIATIONS AND P-VALUE) OVER THE 5 FOLDS. P-VALUE IS CALCULATED BASED ON THE WILCOXON RANK SUM TEST FOR THE PERFORMANCE METRIC OF EACH FOLD BETWEEN CNN-FENN-HHO+ AND OTHER THREE METHODS.

| Methodology | Network structure | Accuracy (%) | Sensitivity (%) | Specificity (%) |
|---|---|---|---|---|
| CNN-MLP | CNN+MLP | 72.58 (3.65, 0.03) | 73.17 (3.16, 0.03) | 67.31 (4.069, 0.03) |
| CNN-ENN | CNN+ENN | 74.92 (3.43, 0.30) | 75.75 (3.03, 0.30) | 70.26 (3.97, 0.30) |
| CNN-FENN | CNN+FENN | 76.37 (2.70, 0.54 | 77.15 (2.78, 0.54) | 72.59 (3.24, 0.54) |
| **CNN-FENN-HHO+** | **CNN+FENN** | **78.25 (2.61, -)** | **79.06 (2.54, -)** | **73.83 (2.86, -)** |

### *B. Application of the proposed FENN and HHO+ on the classification of pulmonary arterial wedge pressure*

Four 18-layer deep learning methods called CNN-MLP, CNN-ENN, CNN-FENN and CNN-FENN-HHO+ for classification of abnormal and normal PAWP using pump flow were compared in TABLE II. This table presents the average and standard deviation of the classification results of PAWP across 5-fold cross validation. The results show that the proposed CNN-FENN-HHO+ improved accuracy, sensitivity and specificity, and outperforms CNN-MLP, CNN-ENN and CNN-FENN methods. Table II also presents the p-values (95% confidence level) based on the Wilcoxon rank sum test to show the statistical significance between the methods. The results show that the proposed CNN-FENN-HHO+ method statistically outperforms the CNN-MLP method.

## IV. DISCUSSION

In this work, an improved HHO (HHO+) algorithm was proposed by combination of HHO, IEEA and QOBL algorithms, being validated on 24 unimodal and multimodal benchmark functions. Although the combination of IEEA, QOBL and HHO algorithm increased the computational complexity of the HHO+ algorithm as shown in Table S6, stagnation in HHO can be avoided more efficiently. The proposed IEEA and employed QOBL concepts generate a set of solutions that can improve the convergence rate and global solution of the original HHO algorithm. This is the advantage of the proposed HHO+ in terms of computational time cost since simply increasing the number of iterations cannot improve the performance of the optimization methods as shown in Fig. S6. In this figure, the convergence curves of all the optimization algorithms except HHO+ stay constant after some iterations for most benchmark functions. The average and standard deviation of 0 in the benchmark functions of F1, F3, F7, F8, F11, F13, F21, F22, F23 and F24 indicate the robustness of the proposed HHO+ on both unimodal and multimodal benchmark functions. Overall, the experimental and statistical (Friedman mean rank and Wilcoxon rank tests) analyses from Table S5 and TABLE I showed the superiority of the proposed HHO+ over the classical HHO and other state-of-the-art algorithms to find the global optima. Fig. S7 and Fig. S8 demonstrate that the performance of HHO+ remains consistently superior even in high-dimensional problems, while the performance of other methods is degraded by increasing dimensions.

A novel Elman neural network called FENN was proposed to improve the classification performance for the considered problem. Moreover, the proposed HHO+ was employed to optimize the three hyper-parameters of CNN-FENN called ILR, LRDF and DP for classification of abnormal and normal PAWP. The results shown in TABLE II indicate that the proposed CNN-FENN-HHO+ method improved accuracy, sensitivity and specificity. This improvement supports the hypothesis of this study that connections between two consecutive time points for the output layer, two consecutive time points for the hidden layer, input layer to output layer, and output layer to hidden layer of a recurrent neural network might fully use the long-term, spatial and temporal information of the input, hidden and output variables of the features, and can therefore lead to higher performance. Appropriate features were automatically extracted from the pump flow signal by convolutional filters in these four deep learning methods. Although the training and testing times of CNN-FENN-HHO+ shown in Tables S7 demonstrate the highest computational complexity compared to other algorithms due to the additional connections between the neurons in the FENN, the proposed algorithm takes only about 1 ms for each heartbeat in the test dataset, which is rapid enough for real-time implementation of the method.

An imbalanced clinical dataset was used in this study for classification of abnormal and normal PAWP. An imbalanced training dataset can impede the convergence of the training process by biasing the training algorithm toward the majority class. Furthermore, it may affect the generalizability of the proposed methods and the performance on the test dataset. Although data augmentation strategies such as oversampling and down-sampling can balance the samples of the classes, they are prone to cause overfitting or losing useful information [76], [77]. Therefore, in this work, the weighted loss function was proposed to overcome the problem.

Lai et al. [11] employed multiple linear regression to estimate PAWP using the slope of the HVAD flow waveform at diastole for HF patients. However, the method resulted in poor $r^2$ of 0.54. Furthermore, the ability of the slope of the HVAD flow waveform at diastole to discriminate PAWP ≥ 20 mmHg (high PAWP) was evaluated by sensitivity and specificity. The results showed a sensitivity of 77% and specificity of 86% for identifying high PAWP. However, in this study, both low and high PAWP were set to abnormal PAWP and the rest as normal. In addition, it is not clear how much data was employed for training, validation, and testing which can change the results significantly.

Estimation of high PAWP (PAWP ≥ 18 mmHg) using the slope of the ventricular filling phase based on 15 HVAD patients was proposed by Grinstein [78]. Although here, too, it is not clear how much data was employed for training, validation and testing, the results derived from multivariable regression demonstrated a sensitivity of 87% and specificity of 95% for only high PAWP. However, in this work, both low and high PAWP was set to abnormal PAWP and 5-fold cross-validation on 25 HVAD patients was used to assess the results. Imamura et al. [79] employed the slope of the ventricular filling phase to estimate high PAWP (PAWP ≥ 18 mmHg) using a linear regression. The results show a sensitivity and specificity of 91.5% and 95.2% to predict only high PAWP, respectively.

Although this work only focuses on the classification of normal and abnormal PAWP using pump flow, it has the potential to be extended to detection of many types of cardiac arrhythmias and real-time monitoring using the ECG signal from the proposed deep learning methods. Moreover, the proposed HHO+ algorithm can be employed to solve real problems with unknown search spaces such as path planning, feature selection, electrical power generation systems, manufacturing processes, image processing, bio-informatics and solar photovoltaic parameter estimation [80].

One of the limitations of this study is that the methods were compared and validated on a small clinical dataset. Therefore, a larger clinical dataset is needed to train and validate the proposed methods. Furthermore, the computational complexity of the optimization and training of the CNN-FENN-HHO+ is higher than the other methods. However, the optimization and training of the CNN-FENN-HHO+ method are performed offline, so it does to some extent obviate this issue.

The accuracy, sensitivity and specificity of the proposed method can still be improved by using a more sophisticated neural network for mapping the final features derived by CNN filters to the labeled classes. Furthermore, the performance of deep networks may be improved by using larger training datasets. The relatively small clinical dataset used in this study may have limited the performance of the compared network. Accordingly, future work will include testing other neural network architectures and collecting additional clinical data for training and validating them on independent datasets.

## V. Conclusion

A novel improved HHO (HHO+) optimization algorithm was proposed, which was validated for optimization of 24 unimodal and multimodal benchmark functions. The results showed that the proposed HHO+ can further improve the global solution and convergence rate. Furthermore, a novel Elman neural network called FENN was proposed to improve the performance metrics of the classification problem. The proposed HHO+ was employed to optimize the three hyper-parameters of CNN-FENN called ILR, LRDF and DP. Moreover, four 18-layer deep learning methods called CNN-MLP, CNN-ENN, CNN-FENN and CNN-FENN-HHO+ for classification of abnormal and normal PAWP using pump flow were developed and compared. The proposed methods were evaluated on an imbalanced clinical dataset using 5-fold cross-validation. Useful features were automatically extracted from the low-pass filtered pump flow signal by convolutional filters. The results demonstrated that the combination of CNN, FENN and HHO+ for classification of abnormal and normal PAWP improved the classification performance. The classification of PAWP based on the pump flow signal can be used to adjust the pump speed in real-time and therefore reduce the likelihood of hazardous events like pulmonary congestion and ventricular suction for HF patients. Furthermore, the proposed method can be implemented on a mobile software platform and report the identified abnormal cases to the hospital, which can diminish the mortality rate of patients with HF.

## Acknowledgment

The authors would like to recognize the financial assistance provided by the National Health and Medical Research Council Centers for Research Excellence (APP1079421).